\documentclass{article}

\usepackage{arxiv}

\usepackage[utf8]{inputenc} % allow utf-8 input
\usepackage[T1]{fontenc}    % use 8-bit T1 fonts
\usepackage{hyperref}       % hyperlinks
\usepackage{url}            % simple URL typesetting
\usepackage{booktabs}       % professional-quality tables
\usepackage{amsfonts}       % blackboard math symbols
\usepackage{nicefrac}       % compact symbols for 1/2, etc.
\usepackage{microtype}      % microtypography
\usepackage{lipsum}
\usepackage[dvipsnames]{xcolor}
\usepackage{graphicx}
\usepackage{subcaption}

\title{Do Optimization Methods in Deep Learning Applications Matter?}

\author{
  Buse Melis Ozyildirim\\
  {Department of Computer Science}\\
  Cukurova University\\
  Adana, Turkey \\
  \texttt{mozyildirim@cu.edu.tr} \\
   \And
Mariam Kiran \\
  Energy Sciences Network\\
  Lawrence Berkeley National Laboratory\\
  Berkeley, CA \\
  \texttt{mkiran@es.net} \\
}

\begin{document}
\maketitle

\begin{abstract}
%\lipsum[1]
With advances in deep learning, exponential data growth and increasing model complexity, developing efficient optimization methods are attracting much research attention. Several implementations favor the use of Conjugate Gradient (CG) and Stochastic Gradient Descent (SGD) as being practical and elegant solutions to achieve quick convergence, however, these optimization processes also present many limitations in learning across deep learning applications. Recent research is exploring higher-order optimization functions as better approaches, but these present very complex computational challenges for practical use. Comparing first and higher-order optimization functions, in this paper, our experiments reveal that Levemberg-Marquardt (LM) significantly supersedes optimal convergence but suffers from very large processing time increasing the training complexity of both, classification and reinforcement learning problems. Our experiments compare off-the-shelf optimization functions (CG, SGD, LM and L-BFGS) in standard CIFAR, MNIST, CartPole and FlappyBird experiments.  
The paper presents arguments on which optimization functions to use and further, which functions would benefit from parallelization efforts to improve pretraining time and learning rate convergence.

\end{abstract}

% keywords can be removed

\section{Introduction}
%\lipsum[2]
%\lipsum[3]
With advances in deep learning research, several optimization functions have been put forward and often used as black-box approaches to quickly train models and deploy for use. Many implementations have favored the use of Stochastic Gradient Descent (SGD) due to their simplicity to implement and quick training of models \cite{Bottou_stochasticgradient, lecun-sgd}. Other approaches such as Conjugate Gradient and batch methods, such as Limited-memory Broyden–Fletcher–Goldfarb–Shanno (L-
BFGS), have also been used as stable and easier to train to optimally converge the loss functions. One of the early works in comparing these optimization functions, Le et al. \cite{Le:2011:OMD:3104482.3104516} showed that using L-BFGS greatly increased the accuracy of MNIST classification results compared to SGD and CG approaches. 
Both, SGD and CG are first-order approaches, giving impressive results in both classification and reinforcement challenges \cite{sun2019survey}. Other higher-order approaches such as  quasi-Newton algorithms also provide good solutions but take extremely long to compute due to matrix and line search calculations. For example, L-BFGS, also a quasi-Newton approach, can work with training data of larger batch sizes compared to SGD, however, L-BFGS suffers from restrictive memory requirements preventing it exceed limitations. The Levenberg-Marquardt (LM) approach is better in this case, as it works iteratively and can loop through multiple batches to find the convergence point.

Working with training batch sizes is a complex requirement especially as training data needed for deep learning applications grows. Researchers have compared the effect of batch sizes, training time and prediction accuracy, when using multiple optimization functions \cite{10.1145/3225058.3225135, rafati19, rafati2018deep}. This shows an important observations that the correct optimization functions can impact the prediction accuracy of the machine learning models.

To cope with large data challenges, computational parallelization has often been used to quickly train models and also cope with memory limitations \cite{10.1145/3225058.3225069, NIPS2011_4390, NIPS2015_5761}. Most solutions in the realm of data and model parallelism have focused on improving SGD calculations for the data and model-layer batches, by strategically placing data or code modules on GPU and HPC computing nodes to parallelize computations. Many deep learning libraries such as TensorFlow, PyTorch and Horovod have also been designed to aid in building these parallelizable solutions of SGD and CG functions when building deep learning applications \cite{tensorflow2015-whitepaper, alex2018horovod}. However, we found that have restricted optimization functions available for developers to explore.

In this paper, we conduct experiments to compare optimization methods in two problem classes - deep reinforcement learning and well-known classification applications. Comparing the loss function, computation time and the accuracy results, this paper sheds light on the challenges developers should consider while building their deep learning applications. We explore challenges in terms of (1) how optimizations functions behave in multiple classification and reinforcement learning challenges (2) computational time needed to process these and (3) the effect on the computations of the loss functions. While SGD and CG, are clearly elegant and quick solutions, our experiments reveal that high-order method, LM in particular, is able to find optimal convergence in both classification and reinforcement learning problems. However, working with larger batch sizes, this approach becomes unfeasible.
 
The rest of the paper is arranged as follows: Section 2 presents related work in this area of optimization functions in deep learning research. Section 3 gives the background on classification and reinforcement problems and training their loss functions. Section 3 also details the optimization functions we compare. The experiment methdology is explained Section 4, with details on experiment setups as well as the results obtained. Section 5 discusses these results with the conclusion presented in Section 6.

%Specifically:
%\begin{itemize}
 %   \item We present comparisons on well known classification and reinforcement learning experiments, CIFAR, MNIST, Cartpole and FlappyBird.
  %  \item We are able to show comparisons in computational time and loss function computations across four commonly used optimization functions.
   % \item We present an indepth analysis as to which function is more appropriate to use for future deep learning research.
    
%\end{itemize}
\section{Related Work}

Stochastic Gradient solutions in example classification and deep learning applications have shown to converge at very high-speeds. Particularly when training time and computational resources are expensive, the ease of converging the loss function quickly is particularly an attractive solution \cite{robbins1951, JMLR:v18:16-595}. For example, where comparisons have shown SGD to converge in 9.45 seconds, L-BFGS converges in 151.49 seconds \cite{sgdtolbfgs}. Additionally the iteration cost of SGD is much cheaper than higher-order optimization functions. Additionally, the Conjugate Gradient approach also has shown promise in accelerated convergence and avoids high computational cost of computing Hessian matrix, needed for higher-order methods. Using quadratic problems, it guarantees convergence in $n$ steps where $n$ represents dimension of input \cite{cg}.

Despite of its success, SGD is still difficult to tune and parallelize \cite{Le:2011:OMD:3104482.3104516}. For example, SGD requires extensive tuning of hyperparameters to improve convergence rates. Currently one has to run models with multiple hyperparameters to select the best for their problems. This search makes SGD actually very computational expensive in large training data problems. also being sequential, SGDs are more difficult to parallelize allowing only data parallelism when training the models \cite{10.2307/2004840, doi:10.1137/18M121112X}. In the reinforcement learning space, Rafati et al. \cite{rafati2018deep} notes that deep reinforcement learning applications using SGD require large memory store (Replay Buffers) and the optimization function often gets stuck with local optima, presenting further challenges to model generalization in learning applications. 

Compared to SGD, methods developed using Gauss-Newton can help learn global optima for convex functions and quadratic convergence by means of higher-order derivative functions \cite{levenberg}. These approaches utilizing inverse of Jacobian or Hessian matrices can speed up convergence and find global optima quickly. Conjugate Gradient can produce solutions in $n$ steps for $n$-dimensional unconstrained quadratic problems, it however, requires estimation of Hessian vector products affecting its performance on the estimation approach. Another approach based on the Hessian matrix, L-BFGS, presents memory challenges and can only work with limited data sets. A fourth optimization approach, the Levenberg Marquardt is based on inverse curvature matrix calculation and is efficient for small and medium scaled problems but too expensive for large datasets. LM also computes local minimas of multivariate functions using an iterative approach to solve nonlinear least squares functions. It utilizes both gradient descent and Gauss-Newton approaches.

Batch methods such as L-BFGS or Conjugate Gradient (CG) use a line search procedure, and are often stable to train and easier to converge \cite{Le:2011:OMD:3104482.3104516} These have been demonstrated training in parallel architectures such as GPUs \cite{Raina:2009:LDU:1553374.1553486} and distributed machines \cite{Chu:2006:MML:2976456.2976492}. The main challenge is often with computing the Hessian matrix, especially with relatively large $n$. The L-BFGS and CG use Quasi-Newton methods which only approximate the Hessian matrix without storing the full matrix. Additionally training on large batches can affect the performance of the convergence rate.

Note: In addition to choosing the best optimizer functions, other parameters can also affect the results. These include the number of layers used in neural network, number of epochs, batch sizes, and learning rate, which we will keep constant during our experiments.

\section{Background}
Machine learning algorithms rely on using optimization functions to learn suitable parameters to train their models. The objective is to minimize the loss function between actual and predicted values while training the machine learning models, for high accuracy predictions. 
\subsection{Loss Functions in Machine Learning Problems}
%{\color{red}{-
%Discuss the loss function
%-Convex error global optimum
%-error functions }}

\textbf{Classification Problems.} Machine learning algorithms are optimized by minimizing the error function, and this error function can vary depending on the problem type. For instance, mean squared error function calculates the squared difference between target and prediction values, particular useful in regression problems. It assumes that outputs are real value functions of input with some Gaussian noises. However, for classification problems, this assumption does not hold.

Outputs of classification problems denote the probability of the inputs belonging to positive classes. It is assumed that data is independent and identically distributed, where the cross entropy function is used as error function. The error function for classification problems can be considered as the difference between distributions of correct labels and algorithm outputs. Hence, cross-entropy error function can be defined as minimizing Kullback-Leibler (KL) divergence. In addition to cross-entropy loss, there is another loss type providing maximal margin between classes called hinge loss (L2-regularized). This loss type is penalizing predictions which are incorrect and not confident in. This results, in not only correct predictions, but also high confidence for correct predictions \cite{lossfunctions}. 

\textbf{Reinforcement Learning.} Another class of machine learning problem is reinforcement learning which is formulated with an agent situated in a partially observable environment, learning from past data to make current decisions. The agent receives data in the form of environment snapshots, processed in some manner, with specific relevant features. After receiving information and computing the reward value for future actions in the current state, the agent then acts to change its environment, subsequently receiving feedback on its action in form of rewards, until terminal state is reached. The objective is to maximize the cumulative reward over all actions in the time agent is active \cite{Sutton:2018:RLI:3312046}. In these problems, agents can use Q-learning, as an off-policy learning algorithm, that uses a table to store all Q-values which are state-action combinations, with possible states and action pairs. This table is updated using the Bellman equation, allowing the action to be chosen using a greedy policy, given as $\gamma$ is discounting factor. Better $Q-$values show better chances of getting higher rewards earned at the end of a complete episode. The $Q-$value is calculated using a $Q-$function. It approximates the $Q-$value using prior $Q-$values, a short-term and a discounted future reward. This way to find optimal control policies across all environment states. The problem with this solution is to keep large tables in the memory and processing them. In deep $Q-$-learning, $Q-$value function is approximated with neural network and the loss function is mean squared error between predicted $Q-$ value and the maximum $Q-$ value for next state and reward \cite{deepq, stanfordrl}.

\subsection{Optimization Function Variants}
\subsubsection{Gradient Descent}
Gradient descent is a common optimization algorithm for training neural networks. Based on the idea of updating the tunable parameters to minimize the objective function, us uses the learning rate to converge the loss function. 

Let $L$ be the objective function and $w$ the model parameter. At every timestep $t$ the $w$ is updated based on following equation,
\begin{equation}
    w_{t}=w_{(t-1)}-\alpha\ \partial L/\partial w
\end{equation}

There are two types of gradient descent implementations: stochastic (SGD) and vanilla gradient descent. While vanilla gradient descent updates parameters after processing the complete training dataset, stochastic gradient descent allows to update the parameters sequentially after each training sample. This causes the vanilla gradient descent to converge slowly compared to stochastic gradient descent. Despite of this speed, stochastic gradient descent suffers from fluctuation and causes the learning rate to decrease very slowly. This problem can be resolved using momentum techniques as it uses the past parameter updates to inform the current decisions, thereby reducing the fluctuations. If parameter updates of two sequential timesteps are in similar directions, the momentum increases and decreases the updates. Examples of these include Nesterov Accelerated Gradient, and others using adaptive learning rates such as Adagrad, Adadelta, RMSprop, and Adam. Comparitive studies have revealed that Adadelta, Adagrad and RMSprop approaches provide the best convergence rates \cite{gradientdescent}.

\subsubsection{Conjugate Gradient}

As discussed above, using previous parameter updates is a way to avoid fluctuations. However, we can also use equations to inform when to perform updates as,
\begin{equation}
    \Delta w_{t}=\ -\partial L/(\partial w_{t}\ )+\beta\Delta w_{(t-1)}
\end{equation}

Conjugate gradient works on finding an optimal coefficient $\beta$ to prevent fluctuation. There are several methods for calculating the coefficient $\beta$ in literature such as Fletcher-Rieves, Polak–Ribière, Hestenes-Stiefel, and Dai–Yuan. The formula below shows the $\beta$ calculation of the methods, respectively. 
\begin{equation}
    \beta=\frac{{\partial L/(\partial w_t\ )}^T(\partial L/(\partial w_t\ ))}{\left(\partial L/(\partial w_{t-1}\ )\right)^T(\partial L/(\partial w_{t-1}\ ))}
\end{equation}

\begin{equation}
    \beta=\frac{{\partial L/(\partial w_t\ )}^T\left(\partial L/(\partial w_t\ )-\partial L/(\partial w_{t-1}\ )\right)}{\left(\partial L/(\partial w_{t-1}\ )\right)^T(\partial L/(\partial w_{t-1}\ ))}
\end{equation}

\begin{equation}
    \beta=\frac{{\partial L/(\partial w_t\ )}^T\left(\partial L/(\partial w_t\ )-\partial L/(\partial w_{t-1}\ )\right)}{{(\Delta w_(t-1))}^T\left(\partial L/(\partial w_t\ )-\partial L/(\partial w_{t-1}\ )\right)}
\end{equation}

\begin{equation}
    \beta=\frac{{\partial L/(\partial w_t\ )}^T\partial L/(\partial w_t\ )}{{(\Delta w_(t-1))}^T\left(\partial L/(\partial w_t\ )-\partial L/(\partial w_{t-1}\ )\right)}
\end{equation}

Although CG provides fast convergence, it often results in poor performance \cite{cgnon}. 

\subsubsection{Broyden-Fletcher-Golfarb-Shanno Algorithm}

Quasi-Newton methods approximate the Hessian value to solve unconstrained optimization problems. Broyden-Fletcher-Golfarb-Shanno (BFGS) algorithm is one of the efficient Quasi-Newton approaches \cite{lbfgs}. 

Let $L$ be a continuously twice differentiable loss function, approximation of the inverse Hessian matrix as given below.

\begin{equation}
    H_{t+1} = H_{t}-\frac{H_{t} s_{t} s_{t}^T H_{t}}{s_{t}^T H_{t} s_{t}} + \frac{y_{t} y_{t}^T} {s_{t}^T y_{t}}
\end{equation}

\begin{equation}
    s_{t} = w_{t+1} - w_{t}
\end{equation}

\begin{equation}
    y_{t} = \left(\partial L/\partial w_{t+1}\right) - \left(\partial L/\partial w_{t}\right)
\end{equation}

Parameter update are as following,

\begin{equation}
    \Delta w_t=\-H_{t}^T \left(\partial L/\partial w_{t}\right) 
\end{equation}

\begin{equation}
    w_{t+1} = w_{t} + \alpha \Delta w_t
\end{equation}

\subsubsection{Levenberg - Marquardt Algorithm}
Levenberg - Marquardt (LM) is an optimization method for solving sum of squares of non-linear functions. It is considered a combination of gradient descent and Gauss-Newton method. It starts with gradient descent and as it comes close to the solution, it behaves like a Gauss-Newton method, which it achieves by using a damping factor. While larger damping factor results in gradient descent, small damping factor lead to Gauss-Newton method. 

Let $d$ be a target value and $y$ be the output of the fitting function, $L$ is defined as:
\begin{equation}
    L = \sum_{j=1}^{m} \left(d_{j} - y_{j} \right)^2
\end{equation}

According to the gradient descent approach the update is as below:

\begin{equation}
    \Delta w_t=\alpha J^T \left(d-y\right)
\end{equation}
where $J$ denotes Jacobian matrix of $\frac{\partial y}{\partial w_t }$

According to the Gauss-Newton approach the update is as,

\begin{equation}
    \Delta w_t= \left(J J^T\right)^{-1} J^T \left(d-y\right)
\end{equation}

Since LM combines these methods, the update rule is,

\begin{equation}
    \Delta w_t= \left(J J^T + \lambda I \right)^{-1} J^T \left(d-y\right)
\end{equation}

where $I$ denotes the identity matrix and $\lambda$ represents the damping factor. If the new parameter values result in lower errors than previous ones, new parameter values are accepted and $\lambda$ is decreased. Otherwise, the new parameter set is rejected and $\lambda$ value is increased \cite{lm}.

\section{Methodology}

\subsection{Datasets and Experiments}

We conducted four experiments - 2 classification and 2 reinforcement learning problems:
\begin{description}

\item[CIFAR.] CIFAR-10 and CIFAR-100 datasets were collected for classification studies by Alex Krizhevsky, Vinod Nair, and Geoffrey Hinton by labelling subsets of 80 million tiny images. CIFAR-10, used in this work, consists of 60000 images with size 32x32 categorized into 10 classes. The data is divided into five training batches and one test batch. Each class in the training batches has 5000 images \cite{cifar}.

\item[MNIST.] MNIST is a handwritten digits dataset including 60000 training and 10000 test samples. It is subset of the NIST dataset. Each image is bilevel and has size of 28x28. In this study, we used the training phase images as normalized data \cite{mnist}.

\item[Flappy Bird.] Flappy bird is a single player video game where the player directs the bird through space between pipes. There are two actions for each state. If the player presses `Up', the bird orients upward, if none the bird descends at a constant rate. Although the output of the Flappy bird game is 284x512, due to the memory issues in our experiment, the image is resized to 84x84. Meanwhile each image is converted to 0-255 color range \cite{flappy}.  

\item[OpenAI gym models: CartPole Experiment.] Gym is a library providing various reinforcement environments to develop and compare algorithms. Since it is compatible with different kinds of libraries, it is frequently used as standard environment. It has several categories such as Atari games, classical control problems, robotics, etc. In these experiments, CartPole-v1 environment is chosen from classical control problems for testing \cite{gymdataset}. 

\end{description}

Note: All tests were done on the same machine. The machine has Intel Core i7 CPU, 16 GB memory and NVIDIA GeForce GTX 1070 8 GB GPU. PyTorch was chosen as the implementation library.

\subsection{Experimental Setup: Optimization functions}

Hyperparameters directly affect the performance of optimization algorithms, with example parameters number of layers, kernel sizes, number of kernels, can all affect the performance of convolutional neural network. Other parameters such as learning rate, number of epochs, and batch sizes are all critical parameters of training process. Some optimizers also include line search steps which include maximum number of iterations. 

Table \ref{hyperparameters} shows the hyperparameter values searched for the four experiments. The learning rates (lr) were searched from $1e-8$ to 0.001, batch sizes (bs) were searched from 32 to 1000 and step size was 32. Maximum iteration for line search (mi) was limited to 10. 

\begin{table}[ht]
\caption{Hyperparameter values for optimizers}
\centering
\begin{tabular}{c c c c c}
\hline\hline
Dataset & LBFGS & SGD & ConjGrad & LM \\
\hline
CIFAR & lr = 1e-6, mi = 10, bs =1000  & lr = 0.001, bs = 1000 & lr = 0.001, bs = 1000& lr = 0.001\\
MNIST & lr = 1e-6, mi = 10, bs =64 & lr = 0.001, bs =64 & lr = 0.001, bs =64 & lr = 0.001\\
Flappy Bird & lr = 1e-6, mi = 20, bs = 32 & lr = 1e-6, bs = 32 & lr = 1e-6, bs =32  & lr = 1e-6\\
Cartpole & lr = 1e-6, mi = 10, bs =32 & lr = 1e-6, bs =32 & lr = 1e-6, bs =32 & lr = 1e-6\\
\hline
\end{tabular}
\label{hyperparameters}
\end{table}

In addition to hyperparameters, Table \ref{netparameters} shows the chosen number of layers and kernel sizes. Due to computational costs, small neural network structures were chosen. 

\begin{table}[ht]
\caption{Number of layers and kernel sizes}
\centering
\begin{tabular}{c c}
\hline\hline
Dataset & Number of layers and Kernel Sizes \\
\hline
CIFAR & 2 conv layers with 5x5 kernels, 1 pooling layer, 3 fully connected layers \\
MNIST & 2 conv layers with 5x5 kernels, 1 pooling layer, 2 fully connected layers \\
Flappy Bird & 3 conv layers with 8x8, 4x4 and 3x3 kernel sizes and 2 fully connected layers\\
Cartpole & 2 fully connected layers\\
\hline
\end{tabular}
\label{netparameters}
\end{table}

%\subsection{Parallel training of models}
%graph of multiple nodes (2, 4 8, nodes

%show results of classification on Cifar in table 
\section{Results}
\subsection{Effect on Training Loss Function}

\begin{figure}[!htb]
  \centering
  \begin{subfigure}[b]{0.45\textwidth}
 \includegraphics[width=1\textwidth]{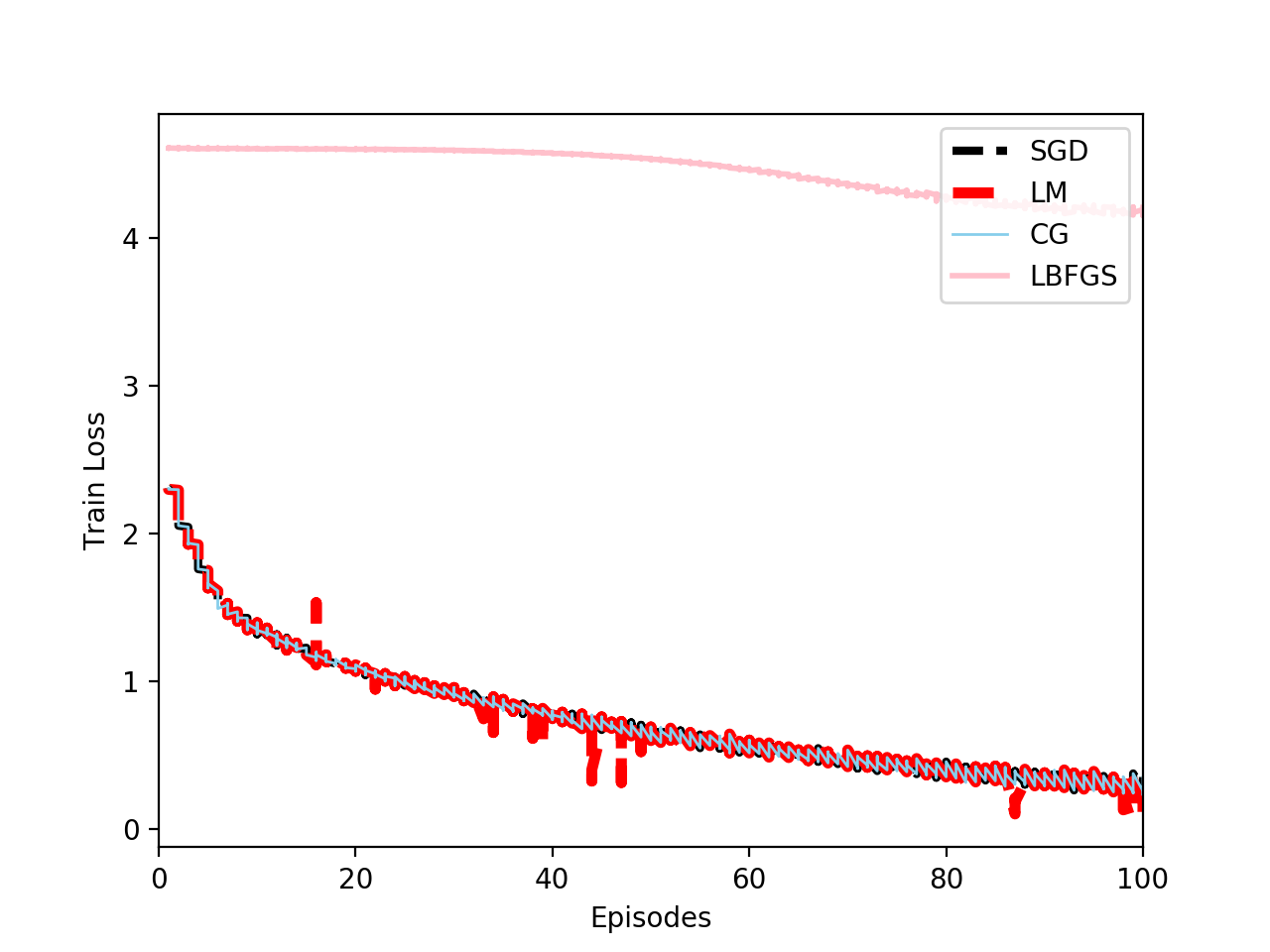}
  \caption{CIFAR.}
  \label{fig:figcifar}
\end{subfigure}
\begin{subfigure}[b]{0.45\textwidth}
\includegraphics[width=1\textwidth]{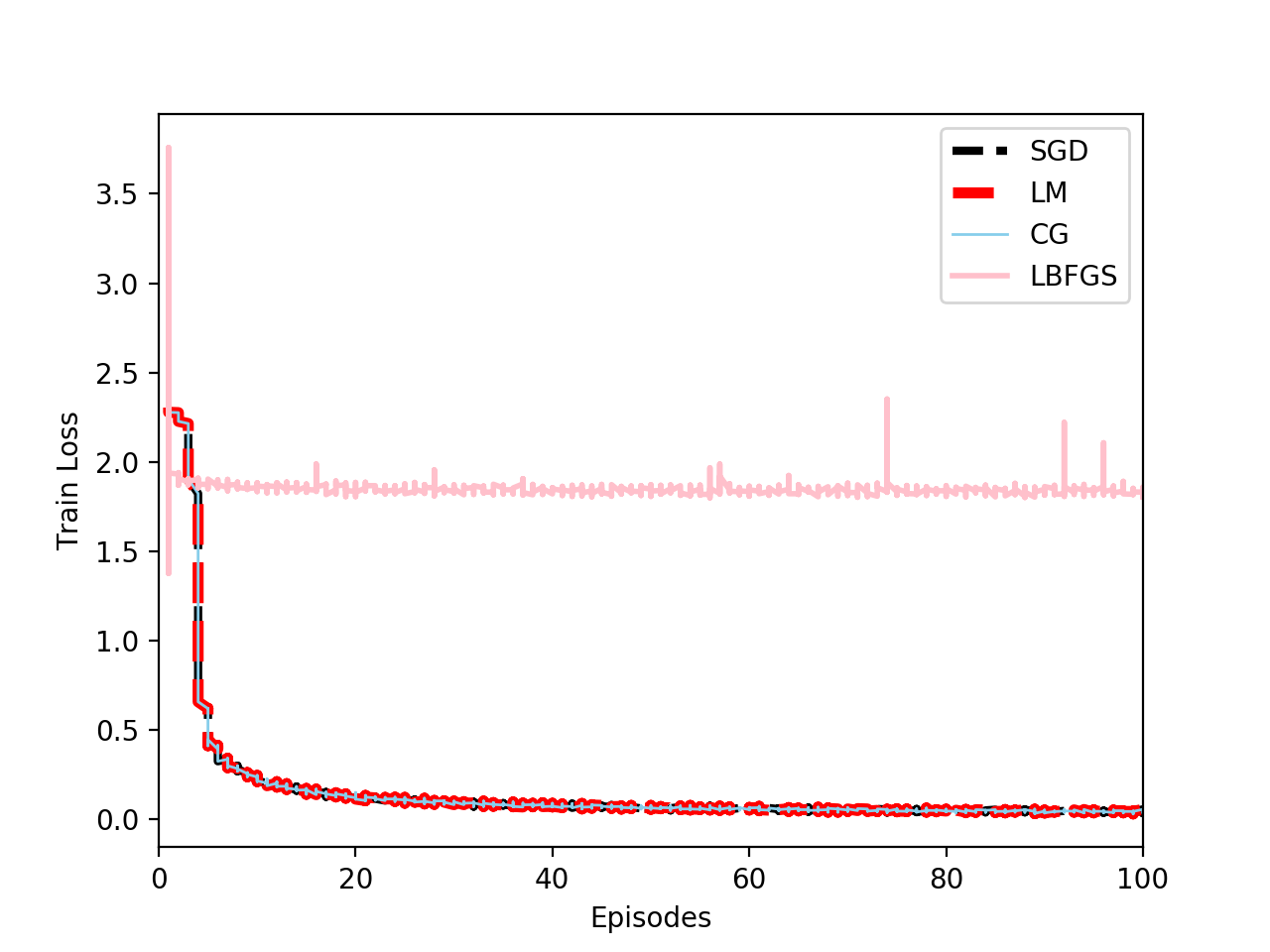}
  \caption{MNIST.}
  \label{fig:figmnist}
\end{subfigure}
 \caption{Training Loss in Classification Experiments.}
 \label{fig:lossclass}
\end{figure}

We compare SGD, CG, L-BFGS and LM for training CNNs in Classification and Reinforcement problems. Figure \ref{fig:lossclass} shows that L-BFGS convergence is much higher than other three optimizers. Additionally, the three optimizers behave similarly to each other. However, LM is able to compute instances of optimal loss in CIFAR, showing that in certain instances it was able to find more optimal results but fluctuated back to previous values due to iterative training approach. 

\begin{figure}[!htb]
  \centering
  \begin{subfigure}[b]{0.45\textwidth}
 \includegraphics[width=1\textwidth]{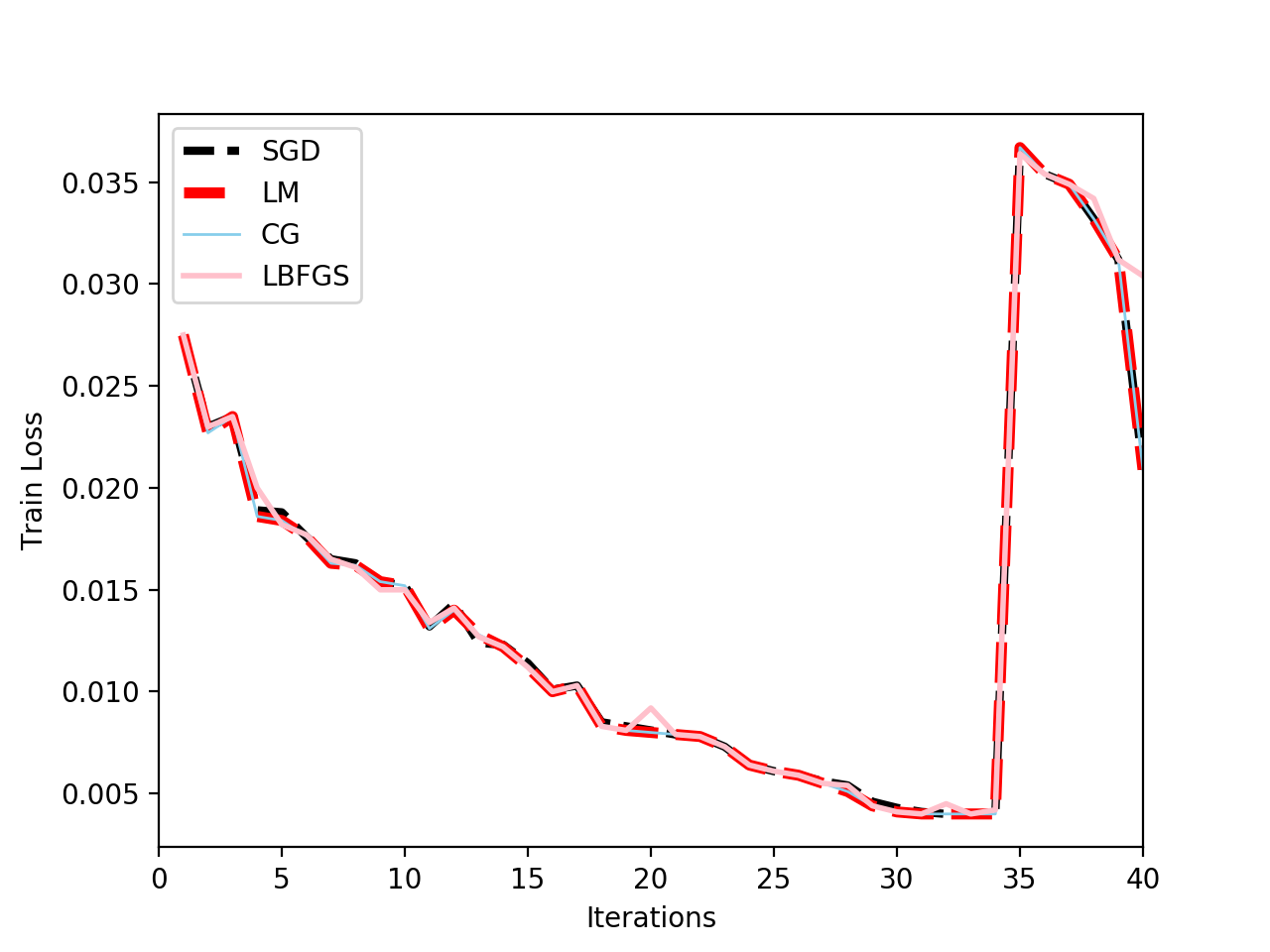}
  \caption{CartPole Episode 1.}
  \label{fig:fig1}
\end{subfigure}
\begin{subfigure}[b]{0.45\textwidth}
\includegraphics[width=1\textwidth]{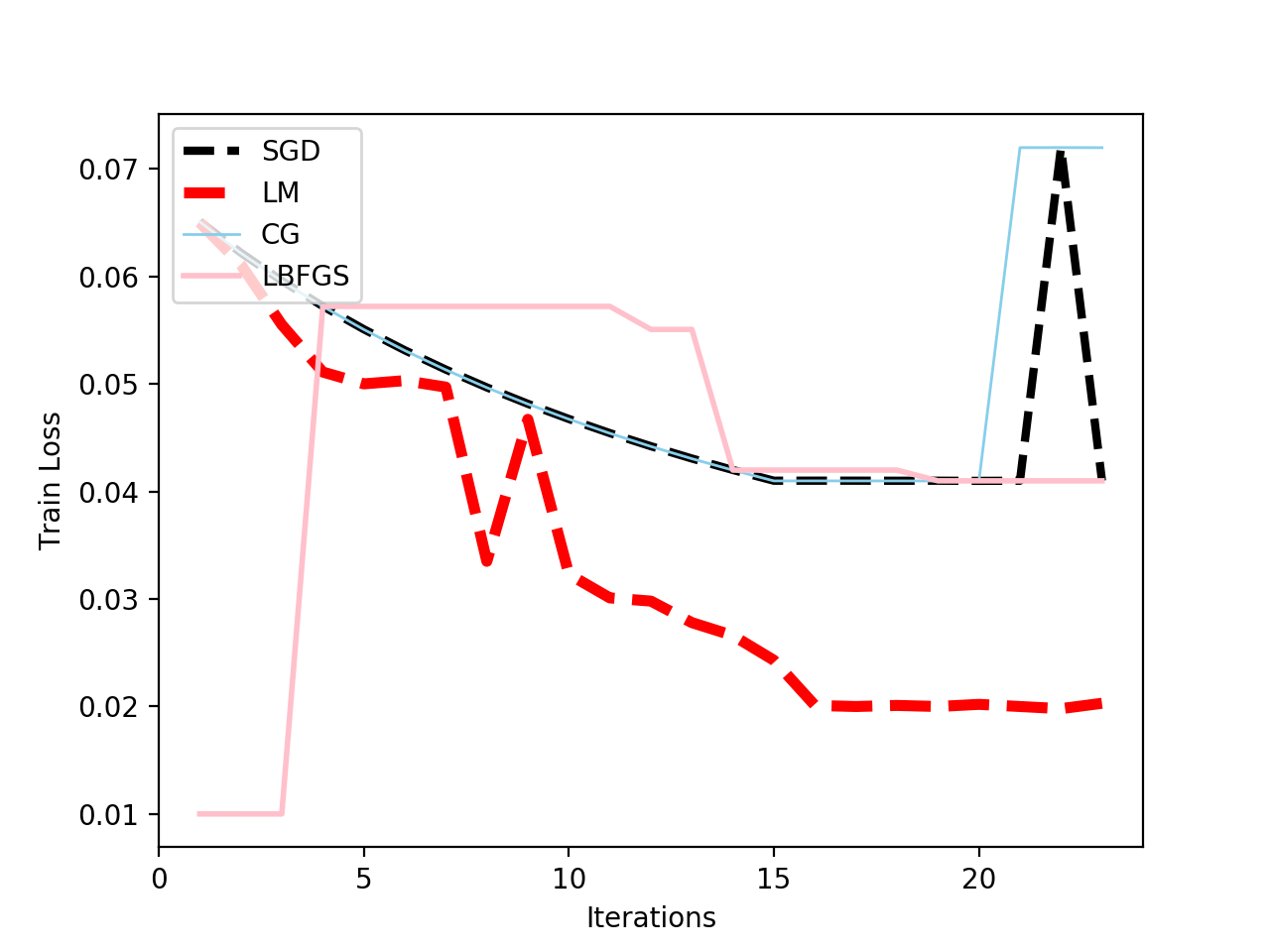}
  \caption{FlappyBird Episode 1.}
  \label{fig:fig1}
\end{subfigure}
 \caption{Training Loss in Deep Reinforcement Experiments.}\label{fig:lossrein}
\end{figure}

Compared to classification, reinforcement loss training shows different behavior (Figure \ref{fig:lossrein}). While the Cartpole follows similar pattern, in FlappyBird, the loss functions are very different. We also observe that LM is able to produce minimal loss computations showing that it can compute better convergence. LM is also better in Flappy Bird as there are only two actions it can perform. While Cartpole has a continual set of actions, we find that when the action class is big, LM has to consider all possibilities and takes longer to converge.

For these results we considered one epoch for Cartpole and Flappy bird, while CIFAR and MNIST results are presented for one iteration. This is because reinforcement learning experiments take longer to compute.

Additionally, we limited the LBFGS to maximum iteration of line search, as it took too long to do line search. LM is able to compute some lower values, but needed to perform additional search to prevent local minima problem.

\subsection{Computation Time}
\begin{figure}[!htb]
  \centering
 \centerline{\includegraphics[width=0.75\textwidth]{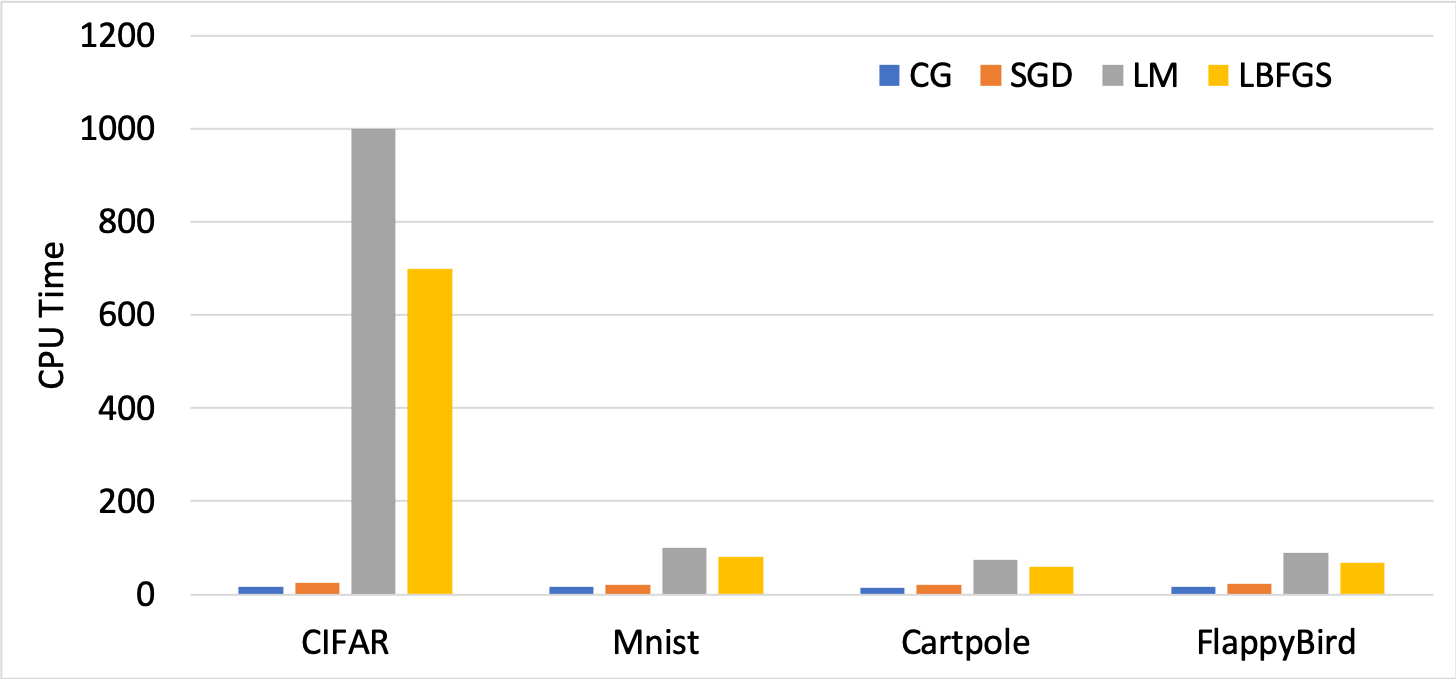}}
  \caption{CPU time for each Optimization Function.}
  \label{fig:counts}
\end{figure}

Figure \ref{fig:counts} compares the computational time per iteration of running the experiments. While we see the that LM was giving optimal convergence, it takes extremely long to converge in all experiments, with the longest time in classification problems such as CIFAR. LM considers all parameters together, hence the higher number of parameters the problem have, the longer it takes to converge.

\subsection{Effect of Q-Learning}

\begin{figure}[!htb]
  \centering
  \begin{subfigure}[b]{0.45\textwidth}
 \includegraphics[width=1\textwidth]{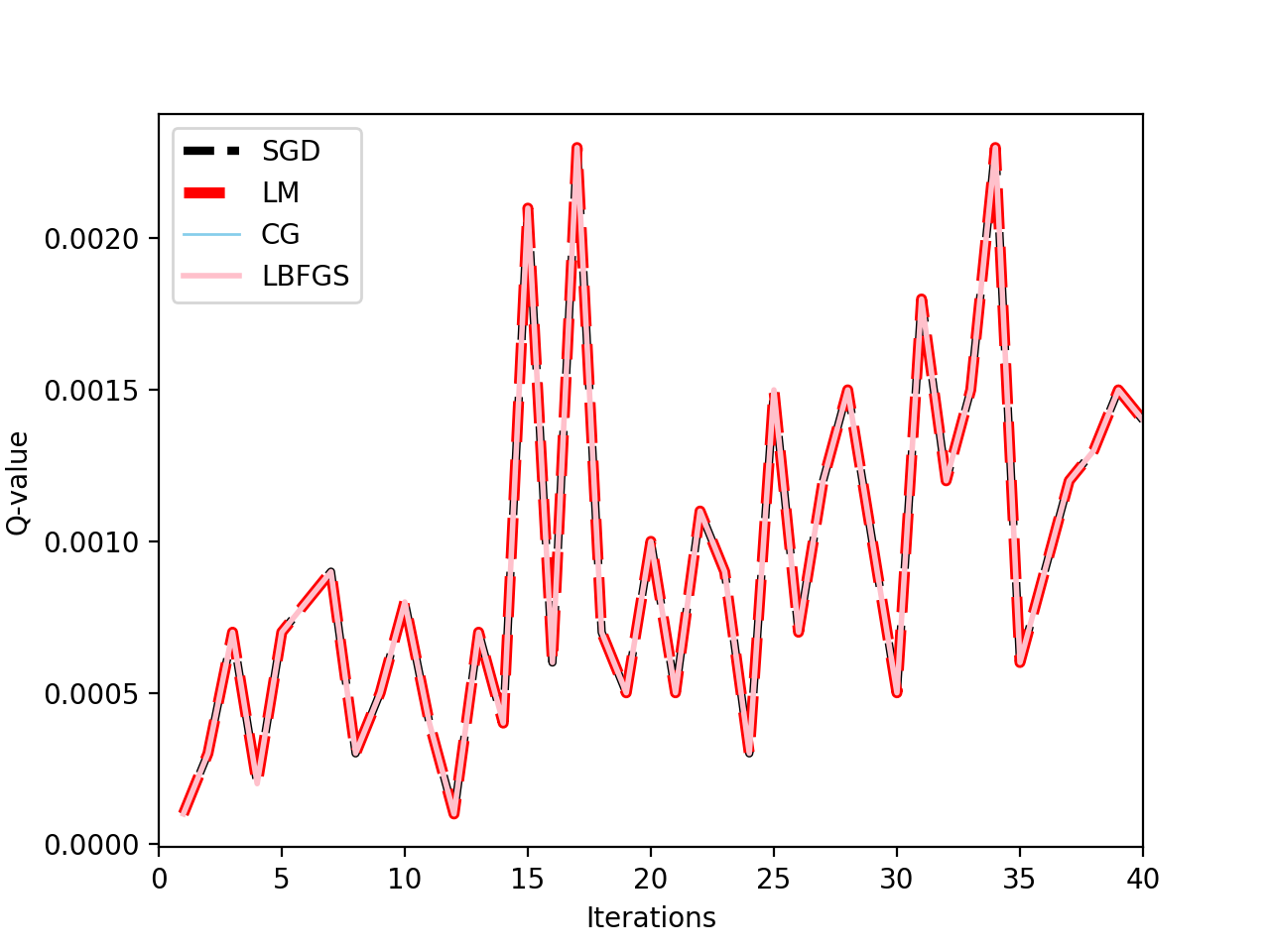}
  \caption{CartPole Episode 1.}
  \label{fig:fig1}
\end{subfigure}
\begin{subfigure}[b]{0.45\textwidth}
\includegraphics[width=1\textwidth]{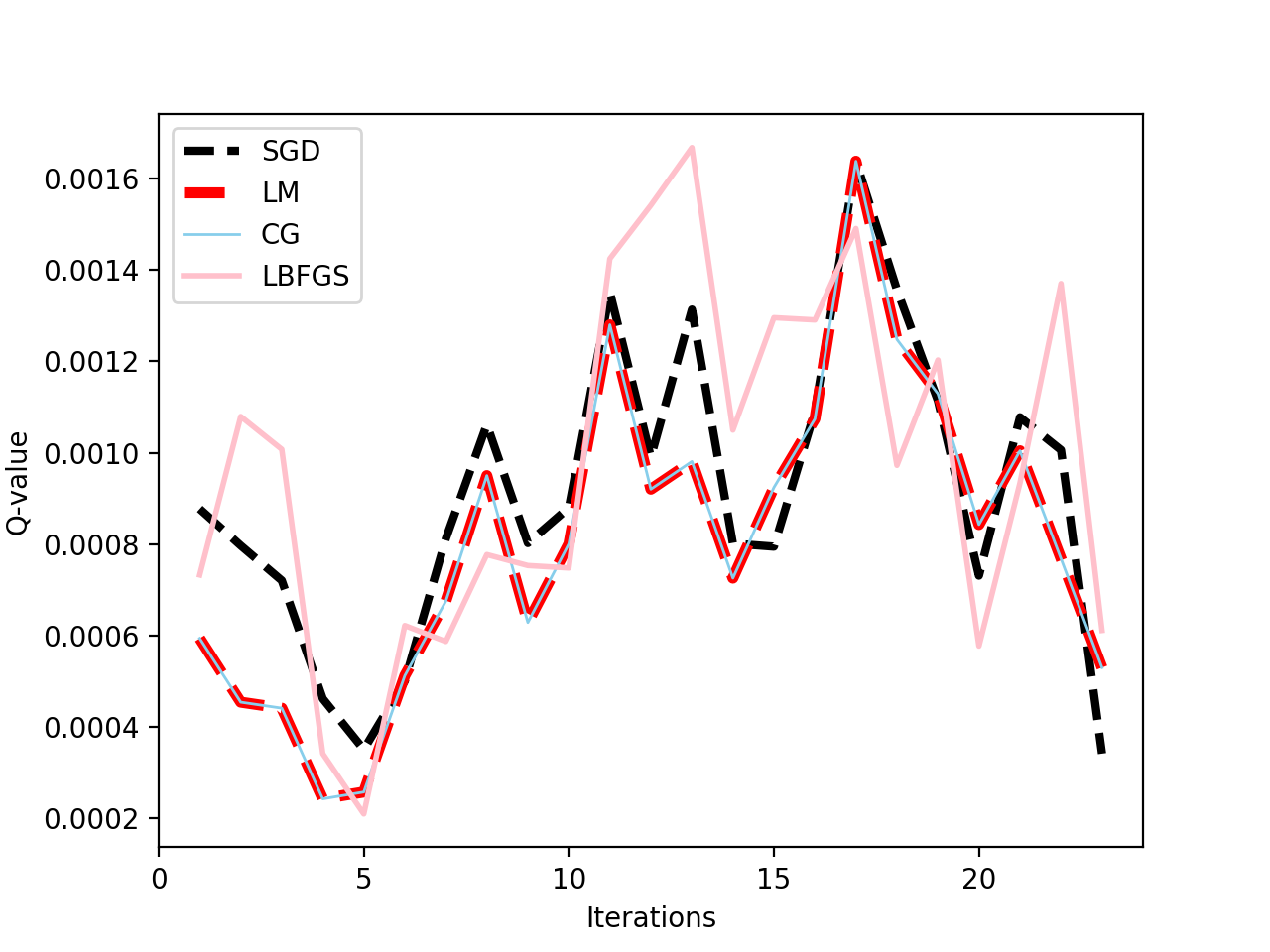}
  \caption{FlappyBird Episode 1.}
  \label{fig:fig1}
\end{subfigure}
 \caption{Q-value in Deep Reinforcement Experiments.}\label{fig:q}
\end{figure}

Figure \ref{fig:q} shows the Q-function value being computed in the reinforcement learning experiments. In Cartpole, we see no effect across all the optimizers, but the flappy bird gives variance across the behaviors. We used the same random seed values to ensure consistency among all experiments. The figure represents one training episode, covering more than one trials. The fluctuations represent new trials for the problems. For Flappy Bird, LM provides better convergence than CG and SGD.But, SGD and CG provides better convergence than LBFGS. This is due to the line search iteration being limited.

\section{Discussion}
\subsection{Comparing the Optimizers}
 Test results show that the second order approaches converge better, but their computational burden is high. Since conjugate gradient descent is between SGD and second order approaches, it has a calculation advantage over second order optimizers.

 Reinforcement learning experiments show that LM performs better if the problem type is regression. Although it provides good results for classification purposes, using mean squared error loss function for classification problems causes computational cost and decrease in performance. Line search algorithm of LBFGS should be chosen properly to take the advantage of second order derivatives. 
 
\subsection{Influence of Training Data Availability}
Deep learning approaches require huge training sets, this causes huge memory need and computational costs. Hence, training step is executed on minibatches. Since training process highly depends on data size, minibatch size should be chosen carefully. While SGD, CG, and LBFGS use minibatches to calculate gradients, LM requires all samples at a time and minibatches cannot be used. While it causes high computation burden, it provides better convergence. 

\subsection{Case for Parallelizing Optimizers}

Second order optimizers such as LBFGS and LM provide high rate convergence if computational cost is ignored. In this study, the line search algorithm of LBFGS was limited to 10 and causes performance drop in LBFGS. On the other hand, although LM provides good results, its computational cost was too high in comparison to other approaches. Using distributed line search algorithm may decrease LBFGS computational time while it increases the convergence rate. In addition to LBFGS, since LM calculates gradient for each sample separately, distributed gradient calculation may make LM faster.

\section{Conclusion}

Machine learning projects consist of five key parts: architecture, data, loss function, optimizers, and hyperparameters. If the one of these are not chosen properly, the algorithm may not perform optimally. One of the challenging steps of machine learning projects is to determine the loss function and optimizers. They should be in accordance with the problem type and amount of data. In this study, performances of different optimizers were tested for two different kinds of machine learning problems on four different datasets with various hyperparameters. Cartpole and Flappy Bird datasets were chosen from reinforcement learning problem and CIFAR and Mnist datasets were chosen from classification problems.

While reinforcement learning uses mean squared loss function, classification problems use cross entropy loss function. The main aim of this study is to investigate the efficiency of optimizers and their appropriate problems. According to the test results, the importance of line search algorithm used in LBFGS is discovered. Line search algorithms brings tradeoff between computational cost and accuracy. The advantage of second order algorithms such as Levenberg-Marquardt, LBFGS is that they provide higher convergence and can represent complex structures. However, high dimensional feature spaces cause high computational costs for these algorithms.

Other disadvantages of Levenberg-Marquardt is that it is a batch algorithm that mini batches cannot be applied. It processes each training sample separately to calculate its gradient. It is useful for mean squared error based problems, classification problems are required to be expressed as regression problems. Conjugate gradient descent approach is accepted as one and half order algorithm and differs from SGD by its direction selection step. In conjugate gradient, rather than calculating gradients before the iteration, direction is determined during the iteration. In comparison to the second order algorithms, it is easier to implement and a little harder than SGD. Results showed that conjugate gradient descent algorithm provided similar results with SGD. 

Finally, second order algorithms provide better convergence under some requirements. While LBFGS requires efficient line search algorithm to provide remarkable results, LM requires low dimension, compact neural networks and parameters to be converged. Conjugate gradient descent algorithm is easy to implement and provides good convergence rates. Since LM calculates gradients per sample separately, if it can be parallelized and computation burden may be reduced. Moreover, the computational burden of line search algorithm used in LBFGS can be updated with parallel version of it.

\bibliographystyle{unsrt}  
\bibliography{references}  

\end{document}